\documentclass[11pt,a4paper]{article}
\usepackage[hyperref]{naaclhlt2019}
\usepackage{times}
\usepackage{latexsym}

\usepackage{url}

\aclfinalcopy 

\usepackage{times}
\usepackage{latexsym}
\usepackage{url}
\usepackage{multirow}
\usepackage{amssymb}
\usepackage{amsmath}
\usepackage{textcomp}
\usepackage{xspace}
\usepackage{array}
\usepackage{subfig}
\usepackage{epsfig}
\usepackage{graphicx}
\usepackage{tablefootnote}
\graphicspath{ {./images/} }

\newcommand\imageclef{\textsc{image-clef}\xspace}
\newcommand\xray{\textsc{x}-ray\xspace}
\newcommand\peir{\textsc{peir}\xspace}
\newcommand\umls{\textsc{umls}\xspace}
\newcommand\mesh{\textsc{mesh}\xspace}
\newcommand\lstm{\textsc{lstm}\xspace}
\newcommand\cnn{\textsc{cnn}\xspace}

\newcommand\bleu{\textsc{bleu}\xspace}
\newcommand\freq{\textsc{frequency}\xspace}
\newcommand\knn{\textsc{nearest-neighbor}\xspace}
\newcommand\imageclefcaption{\textsc{iclef-caption}\xspace}
\newcommand\peirgross{\textsc{peir gross}\xspace}

\newcommand\iuxray{\textsc{iu x-ray}\xspace}

\title{A Survey on Biomedical Image Captioning}

\author{Vasiliki Kougia, John Pavlopoulos, Ion Androutsopoulos
\\
Department of Informatics, Athens University of Economics and Business, Greece \\
\tt \{kouyiav,annis,ion\}@aueb.gr}

\date{}

\begin{document}
\maketitle
\begin{abstract}
Image captioning applied to biomedical images can assist and accelerate the diagnosis process followed by clinicians. This article is the first survey of biomedical image captioning, discussing datasets, evaluation measures, and  state of the art methods. Additionally, we suggest two baselines, a weak and a stronger one; the latter outperforms all current state of the art systems on one of the datasets. 
\end{abstract}

\section{Introduction}
\label{intro}
\begin{figure}[t]
\centering
    \subfloat[General image caption.]{%
        \includegraphics[width=.4\textwidth]{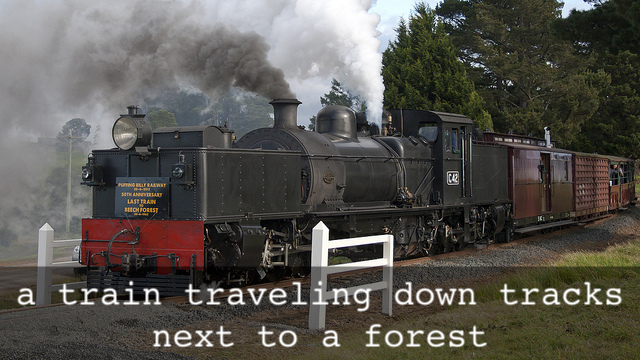}
        \label{fig1:general}
    }\qquad
    \subfloat[Biomedical image caption.]{%
    \includegraphics[width=.4\textwidth]{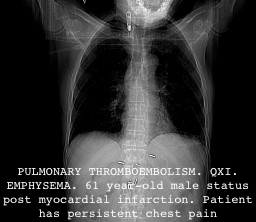}
    \label{fig1:bio}
    }

\caption{Example of a caption produced by the model of \citet{Vinyals2015} for a non-biomedical image (\ref{fig1:general}), and example of a radiology image with its associated caption (\ref{fig1:bio}) from the Pathology Education Informational Resource (\textsc{peir}) Digital Library.}
\end{figure}

Radiologists or other physicians may need to examine many biomedical images daily, e.g.\ \textsc{pet/ct} scans or radiology images, and write their findings as medical reports (Figure~\ref{fig1:bio}). Methods assisting physicians to focus on interesting image regions \cite{Shin2016b} or to describe findings \cite{Jing2018} can reduce medical errors (e.g., suggesting findings to inexperienced physicians) and benefit medical departments by reducing the cost per exam \cite{Bates2001,Lee2017}. 

Despite the importance of biomedical image captioning, related resources are not easily accessible, hindering the emergence of new methods. The publicly available datasets are only three and not always directly available.\footnote{See, for example, \url{http://peir.path.uab.edu/library/} that requires web scrapping.} Also, there is currently no assessment of simple baselines to determine the lower performance boundary and estimate the difficulty of the task. By contrast, complex (typically deep learning) systems are compared to other complex systems, without establishing if they surpass baselines \cite{Zhang2017b,Wang2018}. Furthermore, current evaluation measures are adopted directly from generic image captioning, ignoring the more challenging nature of the biomedical domain \cite{Cohen2014} and thus the potential benefit from employing other measures~\cite{Kilickaya2016}. Addressing these limitations is crucial for the fast development of the field.

This paper is the first overview of biomedical image captioning methods, datasets, and evaluation measures. 
Section~\ref{sect:data} describes publicly available datasets. To increase accessibility and ensure consistent results across systems, we provide code to download and preprocess all the datasets.
Section~\ref{sect:methods} describes biomedical image captioning methods and attempts to compare their results, with the caveat that only two works use the same dataset \cite{Shin2016b,Jing2018} and can be directly compared.
Section~\ref{sect:eval} describes evaluation measures that have been used and introduces two baselines. The first one is based on word frequencies and provides a low performance boundary. The second one is based on image retrieval and the assumption that similar images have similar diagnoses; we show that it is a strong baseline  outperforming the state of the art in at least one dataset. Section~\ref{sect:previous} discusses related (mostly deep learning) biomedical image processing methods for other tasks, such as image classification and segmentation. Section~\ref{sect:conclude} highlights limitations of our work and proposes future directions.

\section{Datasets}
\label{sect:data}

\begin{table*}[ht]
\begin{center}
\begin{tabular}{|l|c|c|c|}
\hline 
\bf Dataset & \bf Images & \bf Tags & \bf Texts \\ \hline
\iuxray & 7,470 \xray{s} & \mesh \& \textsc{mti} extracted terms & 3,955 reports \\ 
\peirgross & 7,442 teaching images & top \textsc{tf-idf} caption words & 7,442 sentences \\
\imageclefcaption & 232,305 medical images & \umls concepts & 232,305 sentences \\
\hline
\end{tabular} 
\\
\vspace*{4mm}
\begin{tabular}{|l|c|c|c|}
\hline 
\bf Dataset & \bf Training Instances & \bf Test Instances & \bf Total \\ \hline
\iuxray & 6,674 & 756 & 7,430 \\ 
\peirgross & 6,698 & 745 & 7,443 \\
\imageclefcaption & 200,074 & 22,231 & 232,305 \\
\hline
\end{tabular}
\end{center}
\caption{Biomedical image captioning publicly available datasets. Images are annotated with tags, which may be medical terms (\iuxray) or words from the captions (\peirgross). A text may be linked to a single image (\peirgross \& \imageclefcaption) or multiple ones (\iuxray). It may comprise a single sentence (\peirgross) or multiple sentences (\imageclefcaption, \iuxray). The lower table shows the number of training and test instances (image-text-tags triples) in each dataset, as used in our experiments. We excluded 40 out of the 7,470 \textsc{iu x-ray} instances, as discussed in the main text.} 
\label{tab:data-bio}
\end{table*}

Datasets for biomedical image captioning comprise medical images and associated texts. Publicly available datasets contain \xray{s} (\iuxray in Table~\ref{tab:data-bio}), clinical photographs (\peirgross in Table~\ref{tab:data-bio}), or a mixture of \xray{s} and photographs (\imageclefcaption in Table~\ref{tab:data-bio}). The associated texts may be single sentences describing the images, or longer medical reports based on the images (e.g., as in Figure~\ref{fig1:bio}). Current publicly available datasets are rather small (\iuxray, \peirgross) or noisy (e.g., \imageclef, which is the largest dataset, was created by automatic means that introduced a lot of noise). We do not include in Table~\ref{tab:data-bio} datasets like the one of \citet{Wang2017}, because their medical reports are not publicly available.\footnote{See, for example, also \url{https://nihcc.app.box.com/v/ChestXray-NIHCC} where only images and text-mined disease labels are released for public use.} Furthermore, we observe that all three publicly available biomedical image captioning datasets suffer from two main shortcomings:
\begin{itemize} 
\item There is a great class imbalance, with most images having no reported findings.
\item The wide range of diseases leads to very scarce occurrences of disease-related terms, making it difficult for models to generalize. 
\end{itemize}

\subsection*{IU X-RAY}
\label{ssec:iuxray}
\citet{DF2015} presented an approach for developing a collection of radiology examinations, including images and narrative reports by radiologists. The authors suggested an accurate anonymization approach for textual radiology reports and provided public access to their dataset through the Open Access Biomedical Image Search Engine (OpenI).\footnote{\url{https://openi.nlm.nih.gov/}} The images are 7,470 frontal and lateral chest \xray{s}, and each radiology report consists of four sections. The `comparison' section contains previous information about the patient (e.g., preceding medical exams); the `indication' section contains symptoms (e.g., hypoxia) or reasons of examination (e.g., age); `findings' lists the radiology observations; and `impression' outlines the final diagnosis. A system would ideally generate the `findings' and `impression' sections, possibly concatenated \cite{Jing2018}. 

The `impression' and `findings' sections of the dataset of \citet{DF2015} were used to manually associate each report with a number of tags (called manual encoding), which were Medical Subject Heading (\mesh)\footnote{ \url{https://goo.gl/iDvwj2}} and RadLex\footnote{\url{http://www.radlex.org/}} terms assigned by two trained coders. Additionally, each report was associated with automatically extracted tags, produced by Medical Text Indexer\footnote{\url{https://ii.nlm.nih.gov/MTI/}} (called \textsc{mti} encoding).
These tags allow systems to learn to initially generate terms describing the image and then use the image along with the generated terms to produce the caption. Hence, this dataset, which is the only one in the field with manually annotated tags, has an added value. From our processing, we found that 104 reports contained no image, 489 were missing `findings', 6 were missing `impression', and 25 were missing both `findings' and `impression'; the 40 image-caption-tags triplets corresponding to the latter 25 reports were discarded in our later experiments. We shuffled the instances of the dataset (image-text-tags triplets) and used 6,674 of them as the training set (images from the 90\% of the reports), with average caption length 38 words and vocabulary size 2,091. Only 2,745 training captions were unique, because 59\% of them were the same in more than one image (e.g., similar images with the same condition). Table~\ref{tab:data-bio} provides more information about the datasets and their splits.

\subsection*{PEIR GROSS}
The Pathology Education Informational Resource (\peir) digital library is a public access image database for use in medical education.\footnote{\url{http://peir.path.uab.edu/library/}} 
Jing et al.\ \shortcite{Jing2018}, who were the first to use images from this database, employed 7,442 teaching images of gross lesions (i.e., visible to the naked eye) from 21 \peir pathology sub-categories, along with their associated captions.\footnote{\peir pathology contains 23 sub-categories, but only 22 contain a gross sub-collection (7,443 images in total). We observe that one image was not included by \citet{Jing2018}.} We developed code that downloads the images for this dataset (called \peirgross) and preprocesses their respective captions, which we release for public use.\footnote{Our code is publicly available at \url{https://github.com/nlpaueb/bio_image_caption}.}
The dataset is split to 6,698 train and 745 test instances (Table~\ref{tab:data-bio}).\footnote{We used 10\% of the dataset for testing, as the 1k images used by Jing et al.\ for validation and testing were not released.} The vocabulary size from the train captions is 4,051 with average caption length 17 words. From the 6,698 train captions only 632 were duplicates (i.e., the same caption for more than one images), which explains why this dataset has a much larger vocabulary than \iuxray, despite the fact that captions are shorter.

\subsection*{ICLEF-CAPTION}
This dataset was released in 2017 for the Image Concept Detection and Caption Prediction (\imageclefcaption) task \cite{ImageCLEF2017} of \imageclef 
\cite{ImageCLEF2018}.
The dataset consists of 184,614 biomedical images and their captions, extracted from  biomedical articles on PubMed Central (\textsc{pmc}).\footnote{\url{https://www.ncbi.nlm.nih.gov/pmc/}} The organizers used an automatic method, based on a biomedical image type hierarchy \cite{Muller2012}, to classify the 5.8M extracted images as clinical or not and also discard compound ones (e.g., images consisting of multiple \xray{s}), but their estimation was that the overall noise in the dataset would be as high as 10\% or 20\% \cite{ImageCLEF2017}.

In 2018, the \imageclefcaption organizers employed a Convolutional Neural Network (\textsc{cnn}), to classify the same 5.8M images based on their type and to extract the non-compound clinical ones, leading to 232,305 images along with their respective captions \cite{ImageCLEF2018}.
Although they reported that compound images were reduced, they noted that noise still exists, with non-clinical images present (e.g., images of maps). Additionally, a wide diversity between the types of the images has been reported \cite{Liang2017}. The length of the captions varies from 1 to 816 words \cite{Su2018,Liang2017}.
Only 1.4\% of the captions are duplicates (associated with more than one image), probably due to the wide image type diversity. The average caption length is 21 words and the vocabulary size is 157,256. A further 10k instances were used for testing in 2018, but they are not publicly available. Hence, in our experiments we split the 235,305 instances into training and test subsets ( Table~\ref{tab:data-bio}).

For tag annotation, the organizers used \textsc{quickumls} \cite{Soldaini2016} to identify concepts of the Unified Medical Language System (\umls) in the caption text, extracting 111,155 unique concepts from the 222,305 captions. Each image is linked to 30 \umls concepts, on average, while fewer than 6k have one or two associated concepts and there are images associated with even thousands of concepts. The organizers observe the existence of noise and note that irrelevant concepts have been extracted, mainly due to the fully automatic extraction process.

\section{Methods} \label{sect:methods}

\citet{Varges2012} employed Natural Language Generation to assist medical professionals turn cardiological findings (e.g., from diagnostic imaging procedures) into fluent and readable textual descriptions. From a different perspective, 
\citet{Schlegl2015} used both the image and the textual report as input to a \textsc{cnn}, trained to classify images with the help of automatically extracted semantic concepts from the textual report. \citet{Kisilev2015a,Kisilev2015b} employed a radiologist to mark an image lesion, and a semi-automatic segmentation approach to define the boundaries of that lesion. Then, they used structured Support Vector Machines \cite{Tsochantaridis2004} to generate semantic tags, originating from a radiology lexicon, 
for each lesion. In subsequent work they used a \textsc{cnn} to rank suspicious regions of diagnostic images and, then, generate tags for the top ranked regions, which can be embedded in diagnostic sentence templates \cite{Kisilev2016}.

\citet{Shin2016b} were the first to apply a \textsc{cnn-rnn} encoder-decoder approach to generate captions from medical images. They used the \textsc{iu x-ray} dataset and a Network in Network \cite{Lin2014} or GoogLeNet \cite{Szegedy2015} as the encoder of the images, obtaining better results with GoogLeNet. The encoder was pretrained to predict (from the images) 17 classes, corresponding to \mesh terms that were frequent in the reports and did not co-occur frequently with other \mesh terms. An \textsc{lstm} \cite{Hochreiter1997} or \textsc{gru} \cite{Cho2014} was used as the \textsc{rnn} decoder to generate image descriptions from the image encodings. In a second training phase, the mean of the \textsc{rnn}’s state vectors (obtained while describing each image) was used as an improved representation of each training image. The original 17 classes that had been used to pretrain the \textsc{cnn} were replaced by 57 finer classes, by applying k-means clustering to the improved vector representations of the training images. The \textsc{cnn} was then retrained to predict the 57 new classes and this led to improved \textsc{bleu} \cite{BLEU2002} scores for the overall \textsc{cnn-rnn} system. The generated descriptions, however, were not evaluated by humans. Furthermore, the generated descriptions were up to 5 words long and looked more like bags of terms (e.g., `aorta thoracic, tortuous, mild'), rather than fluent coherent reports. 

\citet{Zhang2017b} were the first to employ an attention mechanism in biomedical image to text generation, with their  \textsc{mdnet}.\footnote{Zhang et al.\  had introduced earlier TandemNet \cite{Zhang2017a}, which also used attention, but for biomedical image classification. TandemNet could perform captioning, but the authors considered this task as future work, that was addressed with \textsc{mdnet}.} \textsc{mdnet} used \textsc{resnet} \cite{ResNet2016} for image encoding, but extending its skip connections to address vanishing gradients. The image representation acts as the starting hidden state of a decoder \textsc{lstm}, enhanced with an attention mechanism over the image. (During training, this attention mechanism is also employed to detect diagnostic labels.) 
The decoder is cloned to generate a fixed number of sentences, as many as the symptom descriptions. 
This model performed slightly better than a state of the art generic image captioning model \cite{Karpathy2015} in most evaluation measures.

\citet{Jing2018} segment each image to equally sized patches and use \textsc{vgg-19} \cite{VGG2014} to separately encode each patch as a `visual' feature vector. A Multi-Layer Perceptron (\textsc{mlp}) is then fed with the visual feature vectors of each image (representing its patches) and predicts terms from a pre-determined term vocabulary. The word embeddings of the predicted terms of each image are treated as `semantic' feature vectors representing the image. The decoder, which produces the image description, is a hierarchical \textsc{rnn}, consisting of a sentence-level \lstm and a word-level \lstm. The sentence-level \lstm produces a sequence of embeddings, each specifying the information to be expressed by a sentence of the image description (acting as a topic). For each sentence embedding, the word-level \lstm then produces the words of the corresponding sentence, word by word. More precisely, at each one of its time-steps, the sentence-level \lstm of Jing et al.\ examines both the visual and the semantic feature vectors of the image. Following previous work on image captioning, that added attention to encoder-decoder approaches \cite{Xu2015,You2016,Zhang2017b}, an attention mechanism (an \textsc{mlp} fed with the current state of the sentence-level \lstm and each one of the visual feature vectors of the image) assigns attention scores to the visual feature vectors, and the weighted sum of the visual feature vectors (weighted by their attention scores) becomes a visual `context' vector, specifying which patches of the image to express by the next sentence. Another attention mechanism (another \textsc{mlp}) assigns attention scores to the semantic feature vectors (that represent the terms of the image), and the weighted sum of the semantic feature vectors (weighted by attention) becomes the semantic context vector, specifying which terms of the image to express by the next sentence. 
At each time-step, the sentence-level \lstm considers the visual and semantic context vectors, produces a sentence embedding and updates its state, until a stop control instructs it to stop. Given the sentence embedding, the word-level \lstm produces the words of the corresponding sentence, again until a special `stop' token is generated. 
Jing et al.\ showed that their model outperforms models created for general image captioning with visual attention \cite{Vinyals2015,Donahue2015,Xu2015,You2016}.

\begin{table*}[h]
\begin{center}
\begin{tabular}{|l|l|c|c|c|c|c|c|c|}
\hline \bf Method & \bf Dataset & \bf B1 & \bf B2 & \bf B3 & \bf B4 & \bf MET & \bf ROU & \bf CID \bf\\ \hline
Shin et al.~\shortcite{Shin2016b} & \iuxray & 78.5 & 14.4 & 4.7 & 0.0 & - & - & -\\ \hline
\multirow{2}{*}{Jing et al.~\shortcite{Jing2018}} & \iuxray & 51.7 & 38.6 & 30.6 & 24.7 & 21.7 & 44.7 & 32.7 \\
\cline{2-9}
& \peirgross & 30.0 & 21.8 & 16.5 & 11.3 & 14.9 & 27.9 & 32.9 \\\hline
\cline{1-9} 
Wang et al.~\shortcite{Wang2018} & \textsc{chest x-ray 14}$^\dagger$ & 28.6 & 15.9 & 10.3 & 7.3 & 10.7 & 22.6 & - \\ \hline
Zhang et al.~\shortcite{Zhang2017b} & \textsc{bcidr}$^\dagger$ & 91.2 & 82.9 & 75.0 & 67.7 & 39.6 & 70.1 & 2.04 \\ \hline
Gale et al.~\shortcite{Gale2018} & {\small \textsc{frontal pelvic x-rays}}$^\dagger$ & 91.9 & 83.8 & 76.1 & 67.7 & - & - & - \\ \hline
\end{tabular}
\end{center}
\caption{Evaluation of biomedical image captioning methods with \bleu-1/-2/-3/-4 (B1, B2, B3, B4), \textsc{meteor} (\textsc{met}), \textsc{rouge-l} (\textsc{rou}), and \textsc{cider} (\textsc{cid}) percentage scores. \citet{Zhang2017a} and \citet{Han2018} also performed biomedical captioning, but did not provide any evaluation results. Datasets with $\dagger$ are not publicly available; \textsc{bdidr} consists of 1,000 pathological bladder cancer images, each with 5 reports; \textsc{frontal pelvic x-rays} comprises 50,363 images, each supplemented with a radiology report, but simplified to a standard template; \textsc{chest x-ray 14} is publicly available, but without its medical reports.}
\label{tab:bio_i2t} 
\end{table*}

\citet{Wang2018} adopted an approach similar to that of  \citet{Jing2018}, using a \textsc{resnet}-based \cnn to encode the images and an \lstm decoder to produce image descriptions, but their \lstm is flat, as opposed to the hierarchical \lstm of \citet{Jing2018}. Wang et al.\ also demonstrated that it is possible to extract additional image features from the states of the \lstm, much as \citet{Jing2018}, but using a more elaborate attention-based mechanism, combining textual and visual information. 
Wang et al.\ experimented with the same OpenI dataset that Shin et al.\ and Jing et al.\ used. However, they did not provide evaluation results on OpenI and, hence, no direct comparison can be made against the results of Shin et al.\ and Jing et al.\ Nevertheless, focusing on experiments that generated paragraph-sized image descriptions, the results of Wang et al.\ on the (not publicly available) \textsc{chest x-ray} dataset (e.g., \bleu-1 0.2860, \bleu-2 0.1597) are much worse than the OpenI results of Jing et al.\ (e.g., \bleu-1 0.517, \bleu-2 0.386), possibly because of the flat (not hierarchical) \lstm decoder of Wang et al.\footnote{\textsc{chest x-ray} 14 contains 112,120 \xray images with tags (14 disease labels) and medical reports, but only the images and tags (not the reports) are publicly available.} 

\imageclefcaption run successfully for two consecutive years \cite{ImageCLEF2017,ImageCLEF2018} and stopped in 2019. Participating systems (see Table~\ref{tab:imageclef}) used image similarity to retrieve images similar to the one to be described, then aggregating the captions of the retrieved images; or they employed an encoder-decoder architecture; or they simply classified each image based on \umls concepts and then aggregated the respective \umls `semantic groups'\footnote{\url{https://goo.gl/GFbx1d}} to form a caption. 
\citet{Liang2017} used a pre-trained \textsc{vggnet} \cnn encoder and an \lstm decoder, similarly to \citet{Karpathy2015}. They trained three such models on different caption lengths and used an \textsc{svm} classifier to choose the most suitable decoder for the given image. Furthermore, they used a 1-Nearest Neighbor method to retrieve the caption of the most similar image and aggregated it with the generated caption. \citet{Zhang2018}, who achieved the best results in 2018, used the Lucene Image Retrieval software (\textsc{lire}) to retrieve images from the training set and then simply concatenated the captions of the top three retrieved images to obtain the new caption. \citet{Abacha2017} used GoogLeNet to detect \umls concepts and returned the aggregation of their respective \umls semantic groups as a caption. \citet{Su2018} and \citet{Rahman2018} also employed different encoder-decoder architectures.

\begin{table}[h]
\begin{center}
\begin{tabular}{|l|c|c|c|}
\hline \bf Team & \bf Year & \bf Approach & \bf BLEU \\ \hline
Liang et al. &2017& \textsc{ed}+\textsc{ir} & 26.00 \\
Zhang et al. &2018& \textsc{ir} & 25.01 \\
Abacha et al. &2017& \textsc{cls} & 22.47 \\
Su et al. &2018& \textsc{ed} & 17.99 \\
Rahman &2018& \textsc{ed} & 17.25 \\
\hline
\end{tabular}
\end{center}
\caption{Top-5 participating systems at the \imageclefcaption competition, ranked based on average \bleu (\%), the official evaluation measure. Systems used an encoder-decoder (\textsc{ed}), image retrieval (\textsc{ir}), or classified \umls concepts (\textsc{cls}). We exclude 2017 systems employing external resources, which may have seen test data during training \cite{ImageCLEF2017}. 2018 models were limited to use only pre-trained \cnn{s}.}
\label{tab:imageclef}
\end{table}

Gale et al.~\shortcite{Gale2018} argued that existing biomedical image captioning systems fail to produce a satisfactory medical diagnostic report from an image, and to provide evidence for a medical decision. They focused on classifying hip fractures in pelvic \xray{s}, and argued that the diagnostic report of such narrow medical tasks could be simplified to two sentence templates; one for positive cases, including 5 placeholders to be filled by descriptive terms, and a fixed negative one. They used \textsc{densenet} \cite{Huang2017} to get image embeddings and a two-layer \lstm, with attention over the image, to generate the constrained textual report. Their results, shown in Table~\ref{tab:bio_i2t}, are very high, but this is expected due to the extremely simplified and standardized ground truth reports. (Gale et al.\ report an improvement of more than 50 \bleu points when employing this assumption.) The reader is also warned that the results of Table~\ref{tab:bio_i2t} are not directly comparable, since they are obtained from very different datasets.

\section{Evaluation}
\label{sect:eval}
The most common evaluation measures in biomedical image captioning are \bleu \cite{BLEU2002}, \textsc{rouge} \cite{ROUGE2004} and \textsc{meteor} \cite{METEOR2005}, which originate from machine translation and summarization. The more recent \textsc{cider} measure \cite{Vedantam2015}, which was designed for general image captioning \cite{Kilickaya2016}, has been used in only two biomedical image captioning works \cite{Zhang2017b,Jing2018}. \textsc{spice} \cite{Anderson2016}, which was also designed for general image captioning \cite{Kilickaya2016}, has not been used in any biomedical image captioning work we are aware of. Below, we describe each measure separately and discuss its advantages and limitations with respect to biomedical image captioning.

\bleu is the most common measure \cite{BLEU2002}. It measures word n-gram overlap between the generated and the ground truth caption. 
A brevity penalty is added to penalize short generated captions. \bleu-1 considers unigrams (i.e., words), while \bleu-2, -3, -4 consider bigrams, trigrams, and 4-grams respectively. The average of the four variants was used as the official measure in \imageclefcaption. 

\textsc{meteor} \cite{METEOR2005} extended \bleu-1 by employing the harmonic mean of precision and recall (\textsc{f}-score), biased towards recall, and by also employing stemming (Porter stemmer) and synonymy (WordNet). To take into account longer subsequences, it includes a penalty of up to 50\% when no common n-grams exist between the machine-generated description and the reference.

\textsc{rouge-l} \cite{Lin2014} is the ratio of the length of the longest common 
subsequence between the machine-generated description and the reference human description, to the size of the reference  (\textsc{rouge-l} recall); or to the generated description (\textsc{rouge-l} precision); or a combination of the two (\textsc{rouge-l} F-measure). We note that several \textsc{rouge} variants exist, based on different n-gram lengths, stemming, stopword removal, etc., but \textsc{rouge-l} is the most commonly used variant in biomedical image captioning so far.

\textsc{cider} \cite{Vedantam2015} measures the cosine similarity between n-gram \textsc{tf-idf} representations of the two captions (words are also stemmed). This is calculated for unigrams to 4-grams and their average is returned as the final evaluation score. The intuition behind using \textsc{tf-idf} is to reward frequent caption terms while penalizing common ones (e.g., stopwords). However, biomedical image captioning datasets contain many scientific terms (e.g., disease names) that are common across captions (or more generally document collections), which may be mistakenly penalized. 
We also noticed that the scores returned by the provided \textsc{cider} implementation may exceed 100\%.\footnote{We used the official evaluation server implementation \textsc{cider-d} \cite{Capeval2015}.}
We exclude \textsc{cider} results, since these issues need to be investigated further.

\textsc{spice} \cite{Anderson2016} extracts tuples from the two captions (machine-generated, reference), containing objects, attributes and/or relations; e.g., (patient), (has, pain), (male, patient). Precision and recall are computed using WordNet synonym matching between the two sets of tuples, and the \textsc{f1} score is returned. The creators of \textsc{spice} report improved results over both \textsc{meteor} and \textsc{cider}, but it has been noted that results depend on  parsing quality \cite{Kilickaya2016}. When experimenting with the provided implementation\footnote{\url{https://goo.gl/bo11Bz}} of this  measure, we noticed that it failed to parse long texts to evaluate them. Similarly to \textsc{cider}, we exclude \textsc{spice} from further analysis below.

Word Mover's Distance (\textsc{wmd}) \cite{Kusner2015} computes the minimum cumulative cost required to move all word embeddings of one caption to aligned word embeddings of the other caption.\footnote{We used Gensim's implementation of \textsc{wmd} (\url{https://goo.gl/epzecP}) and biomedical word2vec embeddings (\url{https://archive.org/details/pubmed2018_w2v_200D.tar}).} It completely ignores, however, word order, and thus readability, which is one of the main assessment dimensions in the biomedical field \cite{Tsatsaronis2015}. Other previously discussed n-gram based measures also largely ignore word order, but at least consider local order (inside n-grams). \textsc{wmd} scores are included in Table~\ref{tab:baselines} as similarity values $\textsc{wms}=(1+\textsc{wmd})^{-1}$.

\begin{table*}[h]
\begin{center}
\begin{tabular}{|l|l|c|c|c|c|c|c|c|}
\hline \bf Dataset & \bf Baseline & \bf B1 & \bf B2 & \bf B3 & \bf B4 & \bf MET & \bf ROU & \bf WMS \\ \hline
\multirow{3}{*}{\peirgross} & \freq & 29.4 & 6.9 & 0.0 & 0.0 & 12.2 & 17.9 & 23.6\\
\cline{2-9}
& \knn & \textbf{34.6} & \textbf{26.2} & \textbf{20.6} & \textbf{15.6} & \textbf{18.1} & \textbf{34.7} & \textbf{27.5} \\
\cline{2-9} 
& State of the art & 30.0 & 21.8 & 16.5 & 11.3 & 14.9 & 27.9 & -- \\ \hline\hline
\multirow{3}{*}{\iuxray} & \freq & 44.2 & 7.8 & 0.0 & 0.0 & 17.6 & 18.7 & \textbf{30.2}\\
\cline{2-9}
& \knn & 28.1 & 15.2 & 9.1 & 5.7 & 12.5 & 20.9 & 26.0 \\ 
\cline{2-9} 
& State of the art & \textbf{78.5} & \textbf{38.6} & \textbf{30.6} & \textbf{24.7} & \textbf{21.7} & \textbf{44.7} & --  \\ \hline\hline
\multirow{3}{*}{\imageclefcaption} & \freq & \textbf{18.2} & 1.9 & 0.1 & 0.0 & \textbf{4.6} & \textbf{11.1} & \textbf{22.1} \\
\cline{2-9}
& \knn & 7.5 & \textbf{3.0} & \textbf{1.7} & \textbf{1.2} & 4.1 & 8.6 & 20.7 \\ 
\cline{2-9} 
& State of the art & \multicolumn{4}{c|}{26.00} & -- & -- & -- \\ \hline
\end{tabular}
\end{center}
\caption{Evaluation of \freq and \knn on all datasets, with \bleu-1/-2/-3/-4 (B1, B2, B3, B4), \textsc{meteor} (\textsc{met}), \textsc{rouge} (\textsc{rou}), Word Mover's Similarity (\textsc{wms}) percent scores. Best results to date per dataset are also included (state of the art). In \imageclefcaption, only the average \bleu has been reported (26.00).}
\label{tab:baselines} 
\end{table*}

\section {Baselines}
\subsection{Frequency Baseline} 
The first baseline we propose (\freq) uses the frequency of words in the training captions to always generate the same caption. The most frequent word always becomes the first word of the caption, the next most frequent word always becomes the second word of the caption, etc. The number of words in the generated caption is the average length of training captions. Systems should at least outperform this simplistic  baseline and its score should be low across datasets.  

\subsection{Nearest Neighbor Baseline}
The second baseline (\knn) is based on the intuition that similar biomedical images have similar diagnostic captions; this would also explain why image retrieval systems perform well in biomedical image captioning (Table~\ref{tab:imageclef}). We use \textsc{resnet}-18\footnote{\url{https://goo.gl/28K1y2}} to encode images, and cosine similarity to retrieve similar training images. The caption of the most similar retrieved image is returned as the generated caption of a new image. This baseline can be improved by employing an image encoder trained on biomedical images, such as \xray{s} \cite{Rajpurkar2017}. 

\section{Experimental Results}
As shown in Table~\ref{tab:baselines}, \freq scores high when evaluated with \bleu-1 and \textsc{wms}, probably because these measures are based on unigrams. \freq, which simply concatenates the most common words of the training captions, is rewarded every time the most common words appear in the reference captions.

To our surprise, \knn outperforms not only \freq, but also the state of the art in \peirgross, in all evaluation measures (Table~\ref{tab:baselines}). This could be explained by the fact that \peirgross images are phototographs of medical conditions, not \xray{s}, and thus they may be handled better by the  \textsc{resnet}-18 encoder of \knn. In future work, we intend to experiment with an encoder trained on medical images (e.g., \textsc{chexnet}).\footnote{\url{https://stanfordmlgroup.github.io/projects/chexnet/}}

In \iuxray, \knn scores low in all measures, possibly because in this case the images are \xray{s} and the \textsc{resnet}-18 encoder fails to handle them properly. Again, by experimenting with a different encoder, trained on \xray{s}, this baseline might be improved. 

In \imageclefcaption, both of our baselines perform poorly, and much worse than the best system (cf.\ Table~\ref{tab:imageclef}), which achieved average \bleu 26\%. This is partially explained by the size of this dataset (Section~\ref{sect:data}), which contains multiple different images and captions. Moreover, this dataset was created automatically and includes noise and a great diversity of image types (e.g., irrelevant, generic images such as maps) and captions. 

\section{Related Fields}
\label{sect:previous}
Deep learning methods have been widely applied to biomedical images and address various biomedical imaging tasks \cite{Litjens2017}. 
Below, we briefly describe the tasks that are most related to biomedical image captioning, namely biomedical image classification, detection, segmentation, retrieval, as well as general image captioning.

The most related field is image captioning for general images. This is not a new task \cite{Duygulu2002}, but recent work leverages big datasets and has achieved impressive results on generating natural language captions \cite{Karpathy2015}.
The work of Xu et al.~\shortcite{Xu2015} was the first to incorporate attention to the encoder-decoder architecture for image captioning. Appart from improving performance, attention over images helps visualize how the model decides to generate each word and improves interpretability. Image captioning can also be addressed jointly with other tasks, such as video captioning \cite{Donahue2015} or image tagging \cite{Shin2016b}. 

Biomedical image classification aims at classifying a biomedical image as normal or abnormal, or assigning multiple disease labels \cite{Rajpurkar2017,Rajpurkar2018}. Also, it may refer to classifying an abnormality as malignant or benign \cite{Esteva2017}, or assigning other labels (e.g, labels showing the severity of a lesion). A related task is biomedical image detection, which is used to localize and highlight organs or wider anatomical regions \cite{deVos2016} as well as specific abnormalities \cite{Dou2016}. This task is performed as a first step to assist other tasks, such as image classification or segmentation \cite{Bi2017,Rajpurkar2017}.

Biomedical image segmentation aims to divide a biomedical image to different regions representing organs or abnormalities, which can be used for further medical analysis, to learn their features, or classification. The most popular \cnn-based architecture is \textsc{u-net} \cite{Ronnenberger2015}, a version of the network of \citet{Long2015}, altered to produce more precise outputs. Later works \cite{Cicek2016,Milletari2016} extended  \textsc{u-net} for \textsc{3d} image segmentation.

Biomedical image retrieval facilitates searching images in large biomedical databases, based on certain features like symptoms, diseases, and medical cases in general \cite{Liu2016}. Related tasks are also image registration, which performs a spatial alignment of the images \cite{Miao2016,Yang2016}, biomedical image generation \cite{Bahrami2016}, and resolution enhancement of \textsc{2d} and \textsc{3d} biomedical images \cite{Oktay2016}. 

\section{Limitations and Future Work}
\label{sect:conclude}

This paper is a first step towards a more extensive survey of biomedical image captioning methods. We plan to improve it in several ways. Firstly, we hope to investigate to a larger extent the differences between generic image captioning and biomedical image captioning. For example, generic image captioning aims to describe an image, whereas biomedical captioning should ideally help in diagnosis; parts of the image with no diagnostic interest are typically not discussed in a medical report. This investigation may also shed more light to the discussion of appropriate evaluation measures for biomedical image captioning, and the extent to which evaluation measures from generic image captioning, summarizaton, or machine translation are appropriate. 

Secondly, we hope to distill key features from current biomedical image captioning methods (e.g., methods that first tag the images and then generate captions from both the images and their tags vs.\ methods that directly generate captions; methods that retrieve similar images vs.\ methods that do not; types of pretraining used in image encoders and text decoders). This will allow us to provide a more structured and coherent presentation of current methods and highlight possible choices that have not been explored so far.

Thirdly, we plan to consult physicians (e.g., radiologists, nuclear doctors) to obtain a better view of their real-life needs and the degree to which current methods are aligned with their needs. We would also like to contribute to a roadmap of future activities towards integrating biomedical image captioning methods in real-life diagnostic procedures and clinical diagnosis systems. 

\section*{Acknowledgments}
We are grateful to the anonymous reviewers, who suggested several of the future possible improvements mentioned above. We also thank Dr.\ Dimitrios Papamichail for discussions that motivated us to consider biomedical image captioning.

\bibliography{naaclhlt2019}
\bibliographystyle{acl_natbib}

\end{document}